\documentclass[11pt,a4paper]{article}

\usepackage[hyperref]{acl2017}
\usepackage{times}
\usepackage{latexsym}
\usepackage{color}
\usepackage{paralist}
\usepackage{xspace}
\usepackage{amsmath}
\usepackage{amssymb}
\usepackage{pgfplots}
\usepackage{url}

\newcommand{\textnl}[1]{\textsl{`#1'}}

\aclfinalcopy % Uncomment this line for the final submission
\def\aclpaperid{296} %  Enter the acl Paper ID here

\title{Found in Translation:\\
 Reconstructing Phylogenetic Language Trees from Translations}

\author{
\fontsize{11}{10}\selectfont{
\begin{tabular}[t]{c@{\extracolsep{4em}}cc}
Ella Rabinovich$^{\vartriangle\star}$ & Noam Ordan$^{\dagger}$ & Shuly Wintner$^{\star}$ \\
\end{tabular}
}
\\
\fontsize{11}{10}\selectfont{$^{\vartriangle}$IBM Research Haifa, Israel} \\
\fontsize{11}{10}\selectfont{$^{\star}$Department of Computer Science, University of Haifa, Israel} \\
\fontsize{11}{10}\selectfont{$^{\dagger}$The Arab College for Education, Haifa, Israel} \\
\fontsize{10.5}{10}\selectfont{\tt \{ellarabi,noam.ordan\}@gmail.com, \tt shuly@cs.haifa.ac.il} \\
}

\date{}

\begin{document}
\maketitle
\begin{abstract}
Translation has played an important role in trade, law, commerce, politics, and literature for thousands of years. Translators have always tried to be invisible; ideal translations should look as if they were written originally in the target language. We show that traces of the source language remain in the translation product to the extent that it is possible to uncover the history of the source language by looking only at the translation. Specifically, we automatically reconstruct phylogenetic language trees from \emph{monolingual} texts (translated from several source languages). The signal of the source language is so powerful that it is retained even after two phases of translation. This strongly indicates that source language interference is the most dominant characteristic of translated texts, overshadowing the more subtle signals of universal properties of translation.
\end{abstract}

\section{Introduction}
\label{sec:introduction}
Translation has played a major role in human civilization since the rise of law, religion, and trade in multilingual societies. Evidence of scribe translations goes as far back as four millennia ago, to the time of Hammurabi; this practice is also mentioned in the Bible (Esther~1:22;~8:9).  For thousands of years, translators have tried to remain invisible, setting a standard according to which the act of translation should be seamless, and its product should look as if it were written originally in the target language. Cicero ($106$-$43$ BC) commented on his translation ethics, ``I did not hold it necessary to render word for word, but I preserved the general style and force of the language.'' These words were echoed $500$ years later by St.\ Jerome ($347$-$420$ CE), also known as the patron saint of translators, who wrote, ``I render, not word for word, but sense for sense.'' Translator tendency for invisibility has peaked in the past~$150$ years in the English speaking world \citep{venuti2008}, in spite of some calls for ``foreignization'' in translations, e.g., the German Romanticists, especially the translations from Greek by Friedrich H\"{o}lderlin \citep{steiner1975after} and Nabokov's translation of Eugene Onegin. These, however, as both \citet{steiner1975after} and \citet{venuti2008} argue, are the exception to the rule.
In fact, in recent years, the quality of translations has been standardized \href{http://www.iso.org/iso/catalogue_detail.htm?csnumber=59149}{(ISO~$17100$)}. Importantly, the translations we studied in our work conform to this standard.

Despite the continuous efforts of translators, translations are known to feature unique characteristics that set them apart from non-translated texts,  referred to as \emph{originals} here \citep{Toury:1980,Toury:1995,Frawley:1984,Baker:1993}. This is not the result of poor translation, but rather a statistical phenomenon: various features distribute differently in originals than in translations \citep{Gellerstam:1986}.

Several factors may account for the differences between originals and translations; many are classified as \emph{universal} features of translation. Cognitively speaking, all translations, regardless of the source and target language, are susceptible to the same constraints. Therefore, translation products are expected to share similar artifacts. Such universals include \emph{simplification}: the tendency to make complex source structures simpler in the target \citep{Blum-Kulka:1983,Vanderauwera:1985}; \emph{standardization}: the tendency to over-conform to target language standards \citep{Toury:1995}; and \emph{explicitation}: the tendency to render implicit source structures more explicit in the target language \citep{Blum-Kulka:1986,Overas:1990}.

In contrast to translation universals, \emph{interference} reflects the ``fingerprints'' of the source language on the translation product. \citet{Toury:1995} defines interference as ``phenomena pertaining to the make-up of the source text tend to be transferred to the target text''. Interference, by definition, is a language-pair specific phenomenon; isomorphic structures shared by the source and target languages can easily replace one another, thereby manifesting the underlying process of cross-linguistic influence of the source language on the translation outcome.
\citet{pym2008toury} points out that interference is a set of both \emph{segmentational  and  macrostructural  features}.

Our main hypothesis is that, due to interference, languages with shared isomorphic structures are likely to share more features in the target language of a translation. Consequently, the distance between two languages, when assessed using such features, can be retained to some extent in translations from these two languages to a third one. Furthermore, we hypothesize that by extracting structures from translated texts, we can  generate a phylogenetic tree that reflects the ``true'' distances among the source languages. Finally, we conjecture that the quality of such trees will improve when constructed using features that better  correspond to interference phenomena, and will deteriorate using more universal features of translation.

The main contribution of this paper is thus the demonstration that interference phenomena in translation are powerful to an extent that facilitates clustering source languages into families and (partially) reconstructing intra-families ties; so much so, that these results hold even after two rounds of translation. Moreover, we perform analysis of various linguistic phenomena in the source languages, laying out quantitative grounds for the language typology reconstruction results.

\section{Related work}
\label{sec:related-work}
A number of works in historical linguistics have applied methods from the field of bioinformatics, in particular algorithms for generating phylogenetic trees \citep{ringe02IE-Cladistics,nakhleh05JLSAalt,TRPS:TRPS149,Ellison:2006:MLD:1220175.1220210,Boc2010}. Most of them rely on lists of \emph{cognates}, words in multiple languages  with a common origin that share a similar meaning and a similar pronunciation \citep{dyen1992,Rexova2003}. These works all rely on multilingual data, whereas we construct phylogenetic trees from texts in a single language.

The claim that translations exhibit unique properties is well established in translation studies literature \citep{Toury:1980,Frawley:1984,Baker:1993,Toury:1995}. Based on this assumption, several works use text classification techniques employing supervised, and recently also unsupervised, machine learning approaches, to distinguish between originals and translations \citep{Baroni2006alt,Ilisei2010,koppel-ordan:2011:ACL-HLT2011,vered:noam:shuly,TACL618alt,udi:llc}. The features used in these studies reflect both universal and interference-related traits. Along the way, interference was proven to be a robust phenomenon, operating in every single sentence, even on the morpheme level \citep{udi:llc}. Interference can also be studied on pairs of source- and target languages and focus, for example, on word order \citep{eetemadi-toutanova:2014}.

The powerful signal of interference is evident, e.g., by the finding that a classifier trained to distinguish between originals and translations from one language, exhibits lower accuracy when tested on translations from another language, and this accuracy deteriorates proportionally to the distance between the source and target languages \citep{koppel-ordan:2011:ACL-HLT2011}. Consequently, it is possible to accurately distinguish among translations from various source languages \citep{Halteren08}.

A related task, identifying the native tongue of English language students based only on their writing in English, has been the subject of recent interest \citep{tetreault-blanchard-cahill:2013:BEA}.
The relations between this task and identification of the source language of translation has been emphazied, e.g., by \citet{tsvetkov-EtAl:2013:BEA8}.
English texts produced by native speakers of a variety of languages have been used to reconstruct phylogenetic trees, with varying degrees of success \citep{NagataW13,BerzakRK14}. In contrast to language learners, however, translators translate into their mother tongue, so the texts we studied were written by highly competent native speakers. Our work is the first to construct phylogenetic trees from translations.

\section{Methodology}
\label{sec:materials-methodology}
\subsection{Dataset}
\label{sec:dataset}
This corpus-based study uses Europarl \citep{Koehn05Europarl}, the proceedings of the European Parliament and their translations into all the official European Union (EU) languages. Europarl is one of the most popular parallel resources in natural language processing, and has been used extensively in machine translation.
We use a version of Europarl spanning the years 1999 through 2011, in which the direction of translation has been established through a comprehensive \mbox{cross-lingual} validation of the speakers’ original language \citep{ella-shuly:corpus:2015}.

All parliament speeches were translated\footnote{The common practice is that one translates into one's native language; in particular, this practice is strictly imposed in the EU parliament where a translator must have perfect proficiency in the target language, meeting very high standards of accuracy.} from the original language into all other EU languages ($21$ at the time) using English as an intermediate, \textit{pivot} language. We thus refer to translations into English as \textit{direct}, while translations into all other languages, via English as a third language, are \textit{indirect}. We hypothesize that indirect translation will obscure the markers of the original language in the final translation. Nevertheless, we expect (weakened) fingerprints of the source language to be identifiable in the target despite the pivot, presumably resulting in somewhat poorer phylogenetic trees.

We focus on $17$ source languages, grouped into $3$ language families: Germanic, Romance, and Balto-Slavic.\footnote{We excluded source languages with insufficient amounts of data, along with Greek, which is the only representative of the Hellenic family.} These include translations to English and to French from Bulgarian (BG), Czech (CS), Danish (DA), Dutch (NL), English (EN), French (FR), German (DE), Italian (IT), Latvian (LV), Lithuanian (LT), Polish (PL), Portuguese (PT), Romanian (RO), Slovak (SK), Slovenian (SL), Spanish (ES), and Swedish (SV). We also included texts written originally in English and French.

All datasets were split on sentence boundary, cleaned (empty lines removed), tokenized, and annotated for part-of-speech (POS) using the Stanford tools \citep{manning-EtAl:2014:P14-5}. In all the tree reconstruction experiments, we sampled equal-sized chunks from each source language, using as much data as available for all languages. This yielded $27,000$ tokens from translations to English, and $30,000$ tokens from translations into French.

\subsection{Features}
\label{sec:features}
Following standard practice \citep{vered:noam:shuly,TACL618alt}, we represented both original and translated texts as feature vectors, where the choice of features determines the extent to which we expect source-language interference to be present in the translation product. Crucially, the features abstract away from the contents of the texts and focus on their structure, reflecting, among other things, morphological and syntactic patterns. We use the following feature sets:
\begin{inparaenum}
\item The top-1,000 most frequent POS trigrams, reflecting shallow   syntactic structure. %(such as perfect aspect, ``have been --ed'').
\item Function words (FW), words known to reflect grammar of texts in numerous classification tasks, as they include non-content words such as articles, prepositions, etc.\ \citep{koppel-ordan:2011:ACL-HLT2011}.\footnote{For  French we used the list of FW available at \url{https://code.google.com/archive/p/stop-words/}.}
\item Cohesive markers \citep{hinkel2001matters}; these words and phrases are assumed to be over-represented in translated texts, where, for example, an implicit contrast in the original is made explicit in the target text with words such as \textnl{but} or \textnl{however}.\footnote{For French we used \url{http://utilisateurs.linguist.univ-paris-diderot.fr/~croze/D/Lexconn.xml}.}
\end{inparaenum}
Note that the first two feature sets are strongly associated with interference, whereas the third is assumed to be universal and an instance of explicitation. We therefore expect trees based on the first two feature sets to be much better than those based on the third.

\subsection{The Indo-European phylogenetic tree}
\label{sec:ie-gold-tree}
The last few decades produced a large body of research on the evolution of individual languages and language families. While the existence of the Indo-European (IE) family of languages is an established fact, its history and origins are still a matter of much controversy \citep{pereltsvaig2015indo}. Furthermore, the actual sub-groupings of languages within this family are not clear-cut \citep{ringe02IE-Cladistics}. Consequently, algorithms that attempt to reconstruct the IE languages tree face a serious evaluation challenge \citep{ringe02IE-Cladistics, Rexova2003120, nakhleh05JLSAalt}.

To evaluate the quality of the reconstructed trees, we define a metric to accurately assess their distance from the ``true'' tree. The tree that we use as ground truth \citep{0295-5075-81-6-68005} has several advantages. First, it is similar to a well-accepted tree \citep{Gray:2003p112} (which is not insusceptible to criticism \citep{pereltsvaig2015indo}). The differences between the two are mostly irrelevant for the group of languages that we address in this research. Second, it is a binary tree, facilitating comparison with the trees we produce, which are also binary branching. Third, its branches are decorated with the approximate year in which splitting occurred. This provides a way to induce the distance between two languages, modeled as lengths of paths in the tree, based on chronological information.

\begin{figure}[hbt]
\begin{center}
\includegraphics[width=0.5\textwidth]{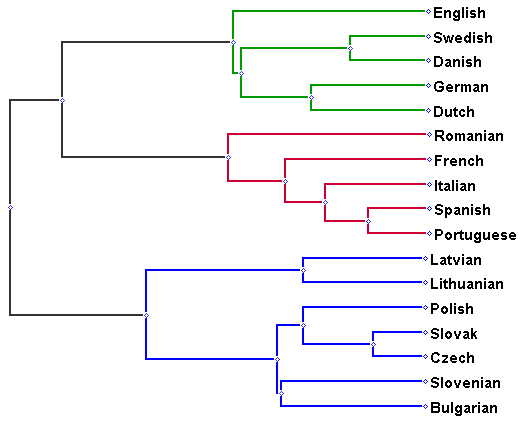}
\end{center}
\caption{Gold standard tree, pruned}
\label{fig:gold-pruned}
\end{figure}

We projected the gold tree \citep{0295-5075-81-6-68005} onto the set of $17$ languages we considered in this work, preserving branch lengths. Figure~\ref{fig:gold-pruned} depicts the resulting gold-standard subtree.

We reconstructed phylogenetic language trees by performing agglomerative (hierarchical) clustering of feature vectors extracted separately from English and French translations. We performed clustering using the variance minimization algorithm \citep{ward1963hierarchical} with Euclidean distance (the implementation  available in the Python SciPy library). All feature values were normalized to a zero-one scale prior to clustering.

\subsection{Evaluation methodology}
\label{sec:eval-methodology}
To evaluate the quality of the trees we generate, we compute their similarity to the gold standard via two metrics: \emph{unweighted}, assessing only structural (topological) similarity, and \emph{weighted}, estimating similarity based on both structure and branching length.

Several methods have been proposed for evaluating the quality of phylogenetic language trees \citep{pompei2011accuracy, wichmann2012quantitative, nouri2016modeling}. A popular metric is the Robinson-Foulds (RF) methodology \citep{robinson1981comparison}, which is based on the symmetric difference in the number of \emph{bi-partitions}, the ways in which an edge can split the leaves of a tree into two sets. The distance between two trees is then defined as the number of splits induced by one of the trees, but not the other.
Despite its popularity, the RF metric has well-known shortcomings; for example, relocating a single leaf can result in a tree maximally distant from the original one \citep{bocker2013generalized}. Additional methodologies for evaluating phylogenetic trees include \emph{branch score distance} \citep{kuhner1994simulation}, enhancing RF with branch lengths, \emph{purity score} \citep{heller2005bayesian}, and \emph{subtree score} \citep{teh2009bayesian}. The latter two ignore branch lengths and only consider structural similarities for evaluation.

We opted for a simple yet powerful adaptation of the L2{-}norm to leaf-pair distance, inherently suitable for both unweighted and weighted evaluation. Given a tree of $N$ leaves, $l_i$, $i\in [1..N]$, the \emph{weighted distance} between two leaves $l_i$, $l_j$ in a tree $\tau$, denoted $D_{\tau}(l_i,l_j)$, is the sum of the weights of all edges on the shortest path between $l_i$ and $l_j$. The \emph{unweighted distance} sums up the \emph{number} of the edges in this path (i.e., all weights are equal to~$1$). The distance $Dist(\tau,g)$ between a generated tree $\tau$ and the gold tree $g$ is then calculated by summing the square differences between all leaf-pair distances (whether weighted or unweighted) in the two trees:

\addtolength{\abovedisplayskip}{-1em}
\begin{equation*}
Dist(\tau,g) = {\sum_{i,j \in [1..N]; i \neq j}(D_{\tau}(l_i,l_j)-D_g(l_i,l_j))^2}
\end{equation*}

\section{Detection of Translations and their Source Language}
\label{sec:detection-of-translationese}
\subsection{Identification of translation}
\label{identification}
We first reconfirmed that originals and translations are easily separable, extending results of supervised classification of \emph{O} vs.\ \emph{T}
(where O refers to original English texts, and T to translated English)
\citep{Baroni2006alt,Halteren08,vered:noam:shuly}
to the $16$ original languages considered in this work. We also conducted similar experiments with French originals and translations. We used $200$ chunks of approximately 2K tokens (respecting sentence boundaries) from both O and T, and normalized the values of lexical features by the number of tokens in each chunk. For classification, we used Platt's sequential minimal optimization algorithm \citep{Keerthi2001,weka} to train support vector machine classifiers with the default linear kernel. We evaluated the results with \mbox{10-fold} \mbox{cross-validation}.

Table~\ref{tbl:class-o-t} presents the classification accuracy of (English and French) O vs.\ T using each feature set. In line with previous works \citep{Ilisei2010,vered:noam:shuly,TACL618alt}, the binary classification results are highly accurate, achieving over $95\%$ accuracy using POS-trigrams and function words for both English and French, and above $85\%$ using cohesive markers.

\begin{table}[hbt]
\centering
%\resizebox{\linewidth}{!}{
\begin{tabular}{lcc}
\multicolumn{1}{c}{\textbf{Feature}} & \textbf{English} & \textbf{French}  \\ \hline
POS-trigrams            & 97.60         & 98.40  \\
Function words          & 96.45         & 95.15  \\
Cohesive markers        & 86.50         & 85.25  \\
\end{tabular}
%}
\caption{Classification accuracy (\%) of English and French O vs. T}
\label{tbl:class-o-t}
\end{table}

\subsection{Identification of source language}
\label{sec:source-language}
Identifying the source language of translated texts is a task in which machines clearly outperform humans \citep{Baroni2006alt}.
%In a 5-way classification of texts translated from Italian, French, Spanish, German, and Finnish, the accuracy was~$92.7\%$ \citep{koppel-ordan:2011:ACL-HLT2011}.
\citet{koppel-ordan:2011:ACL-HLT2011} performed 5-way classification of texts translated from Italian, French, Spanish, German, and Finnish, achieving an accuracy of~$92.7\%$. Furthermore, misclassified instances were more frequently assigned to genetically related languages.

\begin{figure*}[hbt]
\center
\includegraphics[width=1\textwidth]{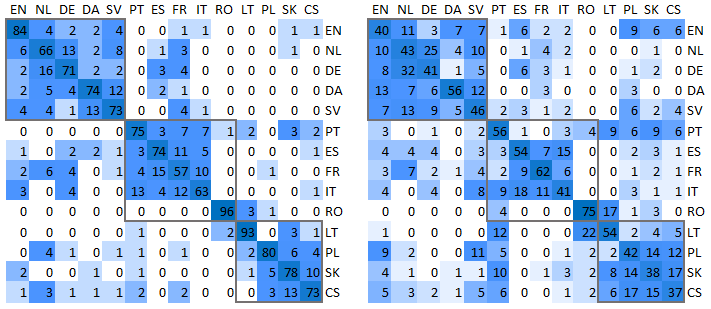}
\caption{Confusion matrix of 14-way classification of English (left) and French (right) translations. The actual class is represented by rows and the predicted one by columns.}
\label{fig:en-fr-conf-matrices}
\end{figure*}

We extended this experiment to $14$ languages representing $3$ language families (the number of languages was limited by the amount of data available). We extracted $100$ chunks of 1,000 tokens each from each source language and classified the translated English (and, separately, French) texts into $14$ classes using the best performing POS-trigrams feature set. \mbox{Cross-validation} evaluation yielded an accuracy of $75.61\%$ on English translations (note that the baseline is $100/14=7.14\%$).

The corresponding confusion matrix, presented in Figure~\ref{fig:en-fr-conf-matrices} (left), reveals interesting phenomena: much of the confusion resides within language families, framed by the bold line in the figure. For example, instances of Germanic languages are almost perfectly classified as Germanic, with only a few chunks assigned to other language families. The evident intra-family linguistic ties exposed by this experiment support the intuition that \mbox{cross-linguistic} transfer in translation is governed by typological properties of the source language. That is, translations from \emph{related} sources tend to resemble each other to a greater extent than translations from more \emph{distant} languages.

This observation is further supported by the evaluation of a three-way classification task, where the goal is to only identify the language family (Germanic, Romance, or Balto-Slavic): the accuracy of this task is~$90.62\%$. Note also that the mis-classified instances of both Romance and Germanic languages are nearly never attributed to Balto-Slavic languages, since Germanic and Romance are much closer to each other than to Balto-Slavic.

Figure~\ref{fig:en-fr-conf-matrices} (right) displays a similar confusion matrix, the only difference being that \emph{French} translations are classified. We attribute the lower \mbox{cross-validation} accuracy ($48.92\%$, reflected also by the lower number of correctly assigned instances on the matrix diagonal, compared to English) to the intervention of the pivot language in the translation process. Nevertheless, the confusion is still mainly constrained to intra-family boundaries.

\section{Reconstruction of Phylogenetic Language Trees}
\label{sec:reconstruction}

\subsection{Reconstructing language typology}
\label{sec:reconstruction-typology}

Inspired by the results reported in Section~\ref{sec:source-language}, we generated phylogenetic language trees from both English and French texts translated from the other European languages. We hypothesized that interference from the source language was present in the translation product to an extent that would facilitate the construction of a tree sufficiently similar to the gold IE tree (Figure~\ref{fig:gold-pruned}).

\begin{figure*}[hbt]
\center
\includegraphics[width=1\textwidth]{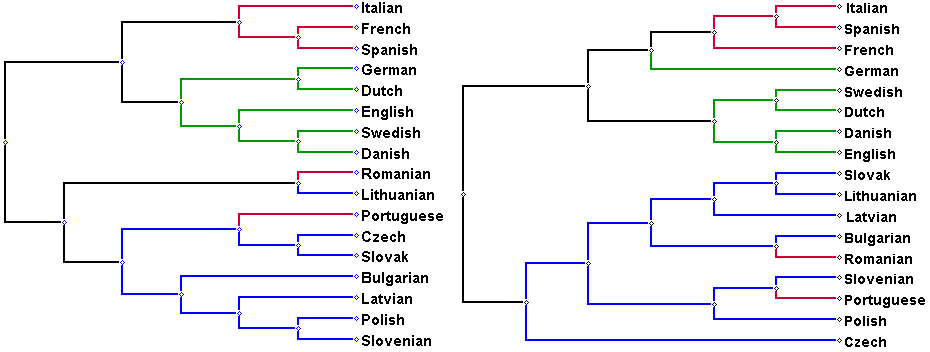}
\caption{Phylogenetic language trees generated with English (left) and French (right) translations}
\label{fig:en-fr-pos-trees}
\end{figure*}

The best trees, those closest to the gold standard, were generated using POS-trigrams: these are the features that are most closely associated with source-language interference (see Section~\ref{sec:features}). Figure~\ref{fig:en-fr-pos-trees} depicts the trees produced from English and French translations using POS-trigrams. Both trees reasonably group individual languages into three language-family branches. In particular, they cluster the Germanic and Romance languages closer than the Balto-Slavic. Capturing the more subtle intra-family ties turned out to be more challenging, although English outperformed its French counterpart on this task by almost perfectly reconstructing the Germanic sub-tree.

We repeated the clustering experiments with various feature sets. For each feature set, we randomly sampled equally-sized subsets of the dataset (translated from each of the source languages), represented the data as feature vectors, generated a tree by clustering the feature vectors, and then computed the weighted and unweighted distances between the generated tree and the gold standard. We repeated this procedure $50$ times for each feature set, and then averaged the resulting distances. We report this average and the standard deviation.\footnote{All the trees, both cladograms (with branches of equal length) and phylograms (with branch lengths proportional to the distance between two nodes), can be found at \url{http://cl.haifa.ac.il/projects/translationese/acl2017_found-in-translation_trees.pdf}}

\subsection{Evaluation results}
\label{sec:eval-results}
The \emph{unweighted} evaluation results are listed in Table~\ref{tbl:evaluation-unweighted}. For comparison, we also present the distance obtained for a random tree, generated by sampling a random distance matrix from the uniform $(0,1)$ distribution. The reported random tree evaluation score is averaged over $1000$ experiments. Similarly, we present \emph{weighted} evaluation results in Table~\ref{tbl:evaluation-weighted}. All distances are normalized to a zero-one scale, where the bounds -- zero and one -- represent the identical and the most distant tree w.r.t. the gold standard, respectively.

\begin{table}[hbt]
\centering
\resizebox{\linewidth}{!}{
\begin{tabular}{lrrrr}
\multicolumn{1}{r}{\textbf{Target language}} & \multicolumn{2}{c}{\textbf{English}} & \multicolumn{2}{c}{\textbf{French}} \\ %\hline
\textbf{Feature}            & AVG   & STD   & AVG   & STD   \\ \hline
POS-trigrams + FW           & 0.362 & 0.07  & \textbf{0.367} & 0.06  \\
POS-trigrams                & \textbf{0.353} & 0.06  & 0.399 & 0.08  \\
Function words              & 0.429 & 0.07  & 0.450 & 0.08  \\
Cohesive markers            & 0.626 & 0.16  & 0.678 & 0.14  \\ %\hline
Random tree                 & 0.724 & 0.07  & 0.724 & 0.07  \\ %\hline
\end{tabular}
}
\caption{Unweighted evaluation of generated trees. AVG represents the average distance of a tree from the gold standard. The lowest distance in a column is boldfaced.}
\label{tbl:evaluation-unweighted}
\end{table}

\begin{table}[hbt]
\centering
\resizebox{\linewidth}{!}{
\begin{tabular}{lrrrr}
\multicolumn{1}{r}{\textbf{Target language}} & \multicolumn{2}{c}{\textbf{English}} & \multicolumn{2}{c}{\textbf{French}} \\ %\hline
\textbf{Feature}            & AVG   & STD   & AVG   & STD   \\ \hline
POS-trigrams + FW           & \textbf{0.278} & 0.03  & \textbf{0.348} & 0.02  \\
POS-trigrams                & 0.301 & 0.03  & 0.351 & 0.03  \\
Function words              & 0.304 & 0.03  & 0.376 & 0.05  \\
Cohesive markers            & 0.598 & 0.12  & 0.636 & 0.07  \\ %\hline
Random tree                 & 0.676 & 0.10  & 0.676 & 0.10  \\ %\hline
\end{tabular}
}
\caption{Weighted evaluation of generated trees. AVG represents the average distance of a tree from the gold standard. The lowest distance in a column is boldfaced.}
\label{tbl:evaluation-weighted}
\end{table}

The results reveal several interesting observations. First, as expected, POS-trigrams induce trees closest to the gold standard among \emph{distinct} feature sets. This corroborates our hypothesis that this feature set carries over interference of the source language to a considerable extent (see Section~\ref{sec:introduction}).
%Furthermore, the lowest standard deviation obtained by POS-trigram-induced trees implies the robustness of these results, since they are clustered more closely around the mean.
Furthermore, function words achieve more moderate results, but still much better than random. This reflects the fact that these features carry over some grammatical constructs of the source language into the translation product.

Finally, in all cases, the least accurate tree, nearly random, is produced by cohesive markers; this is an evidence that this feature is source-language agnostic and reflects the universal effect of explicitation (see Section~\ref{sec:features}). While cohesive markers are a good indicator of translations, they reflect properties that are not indicative of the source language. The combination of POS-trigrams and FW yields the best tree in three out of four cases, implying that these feature sets capture different, complementary aspects of the \mbox{source-language} interference.

Surprisingly, reasonably good trees were also generated from French translations; yet, these trees are systematically worse than their English counterparts. The original signal of the source language is distorted twice: first via a Germanic language (English) and then via a Romance language (French). However, the signal is strong enough to yield a clear phylogenetic tree of the source languages. Interference is thus revealed to be an extremely powerful force, partially resistant to intermediate distortions.

\section{Analysis}
\label{sec:analysis}

\begin{figure*}[hbt]
\begin{center}
\begin{tikzpicture}[font=\small]
  \begin{axis}[
        ybar, axis on top,
        height=4.7cm, width=15.5cm,
        bar width=0.5cm,
        yminorgrids, tick align=inside,
        %ymajorgrids, major grid style={draw=black!10},
        %enlarge y limits={value=.1,upper},
        ymin=0.0, ymax=0.8,
        axis x line*=bottom,
        axis y line*=right,
        y axis line style={opacity=0},
        tickwidth=0pt,
        enlarge x limits=true,
        legend style={
            at={(0.5,1.1)},
            anchor=north,
            legend columns=-1,
            /tikz/every even column/.append style={column sep=0.5cm}
        },
        ylabel={Frequency},
        xticklabels={definite articles\\(per 10 tokens), \textnl{of} constructions\\(per 25 tokens), verb--particle\\(per 250 tokens),
        perfect\\(per 100 tokens), progressive\\(per 500 tokens)},
        xticklabel style={align=center,anchor=base,yshift=-0.75cm},
        xtick=data,
    ]
    \addplot [draw=none, fill=green!70!red] coordinates {
    (0,0.656)(1,0.450)(2,0.702)(3,0.524)(4,0.715)};
    \addplot [draw=none, fill=red!70] coordinates {
    (0,0.734)(1,0.561)(2,0.644)(3,0.412)(4,0.741)};
    \addplot [draw=none, fill=blue!75] coordinates {
    (0,0.754)(1,0.615)(2,0.517)(3,0.383)(4,0.616)};
     \legend{Germanic, Romance, Balto-Slavic}
  \end{axis}
  \end{tikzpicture}
\end{center}
\caption{Frequencies reflecting various linguistic phenomena (Sections~\ref{sec:definite}--~\ref{sec:tense-and-aspect}) in English translations%, averaged  by language family
}
\label{fig:analysis-selected}
\end{figure*}
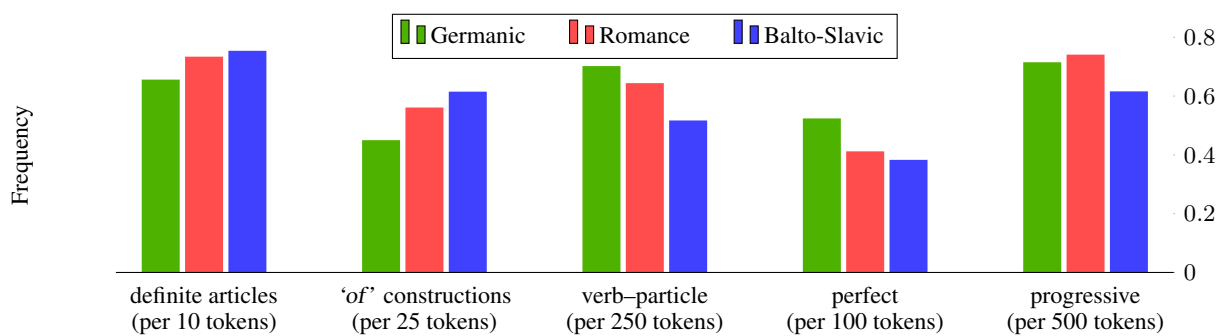

We demonstrated that source-language traces are dominant in translation products to an extent that facilitates reconstruction of the history of the source languages. We now inspect some of these phenomena in more detail to better understand the prominent characteristics of interference. %To this end, we computed several metrics from our dataset; rather than compute metrics for each language individually, we aggregated the data by language family (Germanic, Romance, and Slavic).
For each phenomenon, we computed the frequencies of patterns that reflect it in texts translated to English from each individual language, and averaged the measures over each language family (Germanic, Romance, and Balto-Slavic). Figure~\ref{fig:analysis-selected} depicts the results.

\subsection{Definite articles}
\label{sec:definite}
Languages vary greatly in their use of articles.
%Historically, Proto-Indo-European did not have definite articles, which was also the case in Latin, old English and old German.
Like other Germanic languages, English has both definite (\textnl{a}) and indefinite (\textnl{the}) articles. However, many languages only have definite articles and some only have indefinite articles. Romance languages, and in particular the five Romance languages of our dataset, have definite articles that can sometimes be omitted, but not as commonly as in English. Balto-Slavic languages typically do not have any articles.

Mastering the use of articles in English is notoriously hard, leading to errors in non-native speakers \citep{han2006detecting}. For example, native speakers of Slavic languages tend to \emph{over}use definite articles in German \citep{HirschmannEtAl2013}. Similarly, we expect translations from Balto-Slavic languages to overuse \textnl{the}. We computed the frequencies of \textnl{the} in translations to English from each of the three language families. The results show a significant overuse of \textnl{the} in translations from Balto-Slavic languages, and some overuse in translations from Romance languages.

\subsection{Possessive constructions}
\label{sec:possessive}
Languages also vary in the way they mark possession. English marks it in three ways: with the clitic \textnl{'s} (\textnl{the guest's room}), with a prepositional phrase containing \textnl{of} (\textnl{the room of the guest}), and, like in other Germanic languages, with noun compounds (\textnl{guest room}). Compounds are considerably less frequent in Romance languages \citep{swan2001learner}; Balto-Slavic indicates possession using case-marking.
Languages also vary with respect to whether or not possession is head-marked. In Balto-Slavic languages, the genitive case is head-marked, which reverses the order of the two nouns with respect to the common English \textnl{'s} construction. Since copying word order, if possible across languages, is one of the major features of interference \citep{eetemadi-toutanova:2014}, we anticipated that Balto-Slavic languages will exhibit the highest rate of noun-\textnl{of}-NP constructions. This would be followed by Romance languages, in which this construction is highly common, and then by Germanic languages, where noun compounds can often be copied as such. The results are consistent with our expectations.

\subsection{Verb-particle constructions}
\label{sec:verb-particle}
Verb-particle constructions (e.g., \textnl{turn down}) consist of verbs that combine with a particle to create a new meaning \citep{VPC:2002}. Such constructions are much more common in Germanic languages \citep{iacobini2005verb}, hence we expect to encounter their equivalents in English translations more frequently. We computed the frequencies of these constructions in the data; the results show a clear overuse of verb-particle constructions in translations from Germanic, and an underuse of such constructions in translations from Balto-Slavic.

\subsection{Tense and aspect}
\label{sec:tense-and-aspect}
Tense and aspect are expressed in different ways across languages. English, like other Germanic languages, uses a full system of aspectual distinctions, expressed via perfect and progressive forms (with the auxiliary verbs \textnl{have} or \textnl{be}). Balto-Slavic, in contrast, has no such system, and the distinction is marked lexically, by having two types of verbs. Romance languages are in between, with both lexical and grammatical distinctions.
% (typically through some form of the Latin auxiliary \textnl{hab\={e}re}).
%
We computed the frequencies of perfect forms (defined as the auxiliary \textnl{have} followed by the past participle form), and the progressive forms (defined as the auxiliary \textnl{be} plus a present participle form). Indeed, Germanic overuses the perfect aspect significantly; the use of the progressive aspect also varies across language families, exhibiting the lowest frequency in translations from Balto-Slavic.

\section{Conclusion}
\label{sec:conclusion}
Translations may be considered distortions of the original text, but this distortion is far from random. It depicts a very clear picture, reflecting language typology to the extent that disregarding the sources altogether, a phylogenetic tree can be reconstructed from a monolingual corpus consisting of multiple translations. This holds for the product of highly professional translators, who conform to a common standard, and whose products are edited by native speakers, like themselves. It even holds after two phases of translations.
We are presently trying to extend these results to translations in a different domain (literary texts) into a very different language (Hebrew).

Postulated universals in linguistics \citep{Greenberg-1963} were confronted with much contradicting evidence in recent years \citep{evans:levinson:2009}, and the long quest for translation universals \citep{Universals:2004} should now be viewed in light of our finding: more than anything else, translations are typified by interference.
This does not undermine the force of translation universals:
we demonstrated how explicitation, in the form of cohesive markers, can help identify translations. It may be possible to define classifiers implementing other universal facets of translation, e.g., simplification, which will yield good separation between O and~T.
However, explicitation fails in the reproduction of language typology, whereas interference-based features produce trees of considerable quality.

Remarkably, translations to contemporary English and French capture part of the millennium-old history of the source languages from which the translations were made.
Our trees reflect some of the historical connections among the languages, but of course they are related in other ways, too (whether incidental, areal, etc.). %This may explain why our trees are not  perfect; in particular,
This may explain the case of Romanian in our reconstructed trees: it has been isolated for many years from other Romance languages and was under heavy influence from Balto-Slavic languages.

Very little research has been done in historical linguistics on how translations impact the evolvement of languages. The major trends relate to loan translations \citep{jahr1999language}, or the impact of canonical texts, such as Luther's translation of the Bible to German \citep{russ1994german} or the case of the King James translation to English \citep{crystal2010begat}. It has been attested that for certain languages, up to $30$\% of published materials are mediated through translation \citep{citeulike:675281}. Given the fingerprints left on target language texts, translations very likely play a role in
language change. We leave this as a direction for future research.

\section*{Acknowledgements}We wish to thank the three ACL anonymous reviewers for their constructive feedback. We are grateful to Sergiu Nisioi and Oren Weimann for their advice and helpful suggestions. We are also thankful to Yonatan Belinkov and Michael Katz for insightful and valuable comments.

%\clearpage
\bibliography{final}
\bibliographystyle{acl_natbib}

\end{document}